\documentclass{article}

\usepackage[utf8]{inputenc}
\usepackage[T1]{fontenc}
\usepackage{lmodern}

\usepackage{amsmath,amssymb,amsfonts}

\usepackage{graphicx}
\usepackage{float}
\usepackage{subcaption}
\usepackage{dblfloatfix} 
\graphicspath{{figures/}}

\usepackage{booktabs}
\usepackage{tabularx}
\usepackage{multirow}

\usepackage[dvipsnames]{xcolor}

\usepackage{hyperref}
\usepackage{url}

\usepackage[preprint]{icml2026}

\usepackage{algorithm}
\usepackage{algorithmic}

\usepackage{enumitem}

\usepackage{microtype}
\usepackage{xspace}

\newcommand{\lbone}{\texttt{LAB-Bench}\xspace}
\newcommand{\lbtwo}{\texttt{LABBench2}\xspace}

\icmltitlerunning{\texttt{LABBench2}}

\begin{document}

\twocolumn[
  \icmltitle{\texttt{LABBench2}\\An Improved Benchmark for AI Systems Performing Biology Research}

  \icmlsetsymbol{corr}{$\dagger$}
  \icmlsetsymbol{super}{$\ddagger$}
  \icmlsetsymbol{consult}{$\S$}

  \begin{icmlauthorlist}
    \icmlauthor{Jon M Laurent}{corr,edison}
    \icmlauthor{Albert Bou}{edison}
    \icmlauthor{Michael Pieler}{edison}
    \icmlauthor{Conor Igoe}{edison}
    \icmlauthor{Alex Andonian}{edison}
    \icmlauthor{Siddharth Narayanan}{edison}
    \icmlauthor{James Braza}{edison}
    \icmlauthor{Alexandros Sanchez Vassopoulos}{consult,edison}
    \icmlauthor{Jacob L Steenwyk}{consult,edison}
    \icmlauthor{Blake Lash}{futurehouse,broad}
    \icmlauthor{Andrew D White}{corr,super,futurehouse,edison}
    \icmlauthor{Samuel G Rodriques}{corr,super,futurehouse,edison}
  \end{icmlauthorlist}

  \icmlaffiliation{edison}{Edison Scientific Inc, San Francisco, CA, USA}
  \icmlaffiliation{futurehouse}{FutureHouse, San Francisco, CA, USA}
  \icmlaffiliation{broad}{Broad Institute, Cambridge, MA, USA}

  \icmlcorrespondingauthor{Jon M Laurent}{jon@edisonscientific.com}
  \icmlcorrespondingauthor{Andrew D White}{andrew@edisonscientific.com}
  \icmlcorrespondingauthor{Samuel G Rodriques}{sam@edisonscientific.com}

  \vskip 0.3in
]

\printAffiliationsAndNotice{$^\dagger$Corresponding author. $^\ddagger$Joint technical supervision at Edison Scientific. $^\S$Scientific consultant.}

\begin{abstract}
Optimism for accelerating scientific discovery with AI continues to grow. Current applications of AI in scientific research range from training dedicated foundation models on scientific data to agentic autonomous hypothesis generation systems to AI-driven autonomous labs. The need to measure progress of AI systems in scientific domains correspondingly must not only accelerate, but increasingly shift focus to more real-world capabilities - beyond rote knowledge and even just reasoning to actually measuring the ability to perform meaningful work. Prior work introduced the Language Agent Biology Benchmark (\lbone) as an initial attempt at measuring these abilities. Here we introduce an evolution of that benchmark, \lbtwo, for measuring real-world capabilities of AI systems performing useful scientific tasks. \lbtwo comprises more than 1,900 tasks and is, for the most part, a continuation of \lbone, measuring similar capabilities but in more realistic contexts. We evaluate performance of current frontier models, and show that while abilities measured by \lbone and \lbtwo have improved substantially, \lbtwo provides a meaningful jump in difficulty (model-specific accuracy differences range from $-26\%$ to $-46\%$ across subtasks) and underscores continued room for performance improvement. \lbtwo aims to continue the legacy of \lbone as a de facto benchmark for AI scientific research capabilities and we hope that it continues to help advance development of AI tools for these core research functions. 
To facilitate community use and development, we provide the task dataset at \url{https://huggingface.co/datasets/futurehouse/labbench2} and a public eval harness at \url{https://github.com/EdisonScientific/labbench2}.
\end{abstract}

\section{Introduction}
\label{sec:introduction}

The application of AI for scientific discovery has gained significant momentum in the last two years, with some even proclaiming 2026 the year of AI for Science~\cite{ai_science_2026}. However, science and discovery research are complex endeavors requiring expertise and experience in multiple layers, and measuring the ability of AI systems to operate in these domains remains a daunting task.

The Language Agent Biology Benchmark (\lbone)~\cite{labbench} was the first benchmark for AI systems carrying out practical, real-world biological research tasks, covering literature retrieval, figure and table understanding, lab protocol troubleshooting, database access, and various molecular biology tasks. While it was in many ways ahead of its time, it also necessarily made concessions to make grading easier (using multiple-choice answers), catered to the limitations of LLMs of the time (e.g. DNA sequences provided only in-line rather than use of sequence files or retrieval from sequence databases), and relied on somewhat unrealistic task framing (e.g. pulling information from a provided figure rather than the figure being in the context of a paper.) The capabilities of frontier models have increased substantially since the publication of \lbone, and agentic approaches have grown in popularity and capability as well, leading to multiple examples of near saturation or superhuman performance on \lbone subcategories~\cite{openai_o1_system_card,anthropic_opus45_system_card,paperqa,aviary}. This work introduces an evolution of \lbone called \lbtwo, covering a similar set of task categories while adding a few important new ones, like evaluating quality of literature sources and retrieval from patents and clinical trials. Briefly, our primary contributions are:
\begin{itemize}
    \item We expand the scope and realism of \lbone by introducing \lbtwo, an improved dataset of more than 1,900 tasks spanning literature understanding and retrieval, data access, protocol troubleshooting, molecular biology assistance, and experiment planning.
    \item We extend literature-focused evaluation with new open-response and retrieval-based variants for figures and tables, and add new task families covering patents, clinical trials, and source-quality assessment.
    \item We report baseline results for current frontier models and highlight a substantial increase in difficulty relative to \lbone.
    \item We release the benchmark dataset and provide an evaluation harness for community use.
\end{itemize}

\subsection{Related Work}
\label{sec:related-work}
The original \lbone~\cite{labbench} suite was among the first benchmarks designed to evaluate broad and practical biology-research capabilities in large language models, moving beyond “textbook-style” science questions to tasks that resemble day-to-day research work. Early scientific benchmarks such as SciBench~\cite{scibench}, SciEval~\cite{scieval}, GPQA~\cite{gpqa}, and HLE~\cite{phan2025humanitysexam} provide valuable, scalable measures of scientific knowledge and reasoning. More recently, FrontierScience~\cite{wang_frontierscience_2026} has been released as a step towards i) more realistic reasoning tasks and ii) real-world information-retrieval research tasks. However, these benchmarks are largely constructed as static evaluations that capture rote knowledge rather than the applied workflow that characterizes biomedical research. As AI systems have advanced through improvements in inference scaling, retrieval-augmented approaches, and agentic interaction, benchmark design has increasingly shifted toward higher-fidelity evaluations that test multi-step decision-making and outcomes in research-like settings. 

This includes information retrieval and reasoning benchmarks like BioASQ~\cite{krithara_bioasq-qa_2023}, PubMedQA~\cite{jin_pubmedqa_2019}, and LitQA2 from \lbone~\cite{labbench} for measuring the ability to find and extract important published information, CiteME~\cite{press_citeme_2024} measures the ability of agentic information systems to reliably ground responses in published sources. BrowseComp~\cite{wei_browsecomp_2025} and ~\cite{du_deepresearch_2025} are newer information retrieval benchmarks for the increasingly important deep research-style agent systems that are more effective at multi-source retrieval and reasoning.

More focused scientific domain benchmarks have become increasingly common. DiscoveryBench~\cite{discoverybench} and DISCOVERYWORLD~\cite{jansen_discoveryworld_2024} that measure the ability to make discoveries through application of scientific approaches, BixBench~\cite{bixbench} for assessing proficiency in realistic bioinformatic data analysis scenarios, BioProBench~\cite{bioprobench} and BioLP-Bench~\cite{ivanov_biolp-bench_2024}, BioKGBench~\cite{biokgbench} (biomedical knowledge-graph reasoning), ChemBench~\cite{chembench} (chemistry capabilities), and RxnBench~\cite{rxnbench} (multimodal reaction understanding from scientific literature) exemplify this direction within biology/biomedicine and chemistry. These newer suites typically go deeper within a single subdomain, whereas the LABBench benchmarks aim to cover a somewhat wider span of practical research competencies within one unified framework.

\section{Description}
\label{sec:description}

\lbtwo is composed of more than 1,900 total tasks. Aligning with \lbone, tasks are distributed among five broad categories: Literature retrieval and understanding, Data access, Protocol troubleshooting, Molecular biology assistance, and Experiment planning.

\begin{table}[ht]
\centering
\caption{LABBench2 Task Categories and their counts.$^{*}$Note that SeqQA2 and CloningQA2 additionally have variant modes for sequence delivery: direct prompt injection, file delivery, or retrieval}
\label{tab:task-categories}
\begin{tabular}{llr}
\toprule
\textbf{Category} & \textbf{Variant} & \textbf{Count} \\
\midrule
LitQA3 & & 168 \\
PatentQA & & 121 \\
TrialQA & & 120 \\
SourceQuality & & 150 \\
DbQA2 & & 86 \\
SuppQA2 & & 125 \\
ProtocolQA2 & & 125 \\
SeqQA2 & & 400$^{*}$ \\
CloningQA & & 14$^{*}$ \\
\midrule
\multirow{3}{*}{FigQA2} & retrieve & 101 \\
 & pdf & 101 \\
 & img & 101 \\
\midrule
\multirow{3}{*}{TableQA2} & retrieve & 100 \\
 & pdf & 100 \\
 & img & 100 \\
\midrule
\textbf{Total} & & \textbf{1912} \\
\bottomrule
\end{tabular}
\end{table}

\subsection{Literature understanding}
\label{sec:literature}

\lbone consisted of several subsets covering literature-related tasks, including literature information retrieval (LitQA2), understanding scientific figures and tables (FigQA and TableQA), and retrieving information from supplemental material (SuppQA). For \lbtwo (LB2), updated approaches were developed for these same subtasks, and several new subtask types are added, including evaluating literature source quality and evaluating retrieval of information from patents and clinical trials in addition to research papers.

\paragraph{LitQA3.}
Like LitQA2 in \lbone, LitQA3 in \lbtwo is composed of questions that require retrieval of specific research papers and require reading and reasoning over the full text in order to answer correctly. For LitQA3, the primary improvement is that all questions are open-response. Indeed, a subset of the questions in LitQA3 are ported from LitQA2 and reworded as necessary to create an open-response version, though the majority of questions are brand new.

\paragraph{FigQA2 and TableQA2.}
The original FigQA and TableQA benchmarks presented questions that required nuanced interpretation of information presented in scientific figures and tables in a multiple-choice format. Importantly, in both cases the figure or table in question was presented to the model as an image file and thus measured only the ability to read the image and choose the best option. For the new FigQA2 and TableQA2, all new questions have been written, and we've created three different task variants for each figure or table:

\begin{enumerate}[label=(\roman*)]
    \item \textbf{Image:} Similar to the original Fig/TableQA tasks, but are open-response. Require only reading and understanding the figure or table as an image, which is provided to the model along with the question.
    \item \textbf{Paper:} Rather than the figure image, the entire PDF of the source paper is provided to the model. The correct figure or table containing the information needed to answer must be parsed and interpreted from the distracting context of the full paper.
    \item \textbf{Retrieval:} Here the task is similar to LitQA3, where no context is provided, and the model/agent must retrieve the correct source paper and properly locate the information in the correct figure or table.
\end{enumerate}

\paragraph{PatentQA and TrialQA.}
Accessing and utilizing information contained in other text-based information sources beyond research papers is particularly important for high-value research applications like therapeutic development. Accordingly, \lbtwo contains two benchmark categories assessing retrieval of information from full-text patents and clinical trials. These are framed similarly to LitQA3 tasks, with questions that specifically require a single specific trial or patent be identified unambiguously. Some questions may additionally require reading a table or figure within the patent or trial and are thus multi-modal, similar to the Retrieval variant of FigQA2/TableQA2.

\paragraph{SuppQA2.}
The primary text of most research papers presents a curated and incomplete description of the presented study. Many include extensive supplementary information (SI) to augment the main paper with other important information. SI is highly non-standard and can contain variable types of information including data, detailed experimental methods, or additional findings. \lbone included a sub-set of literature retrieval tasks called SuppQA intended to measure the ability to access these extended information sources, and this is carried forward in \lbtwo with SuppQA2. These are open-response questions that require accessing text, figures, tables, data, etc. from research paper SI files.

\paragraph{SourceQualQA.}
Autonomous scientific discernment is a key capability for AI agents supporting information synthesis. While structured quality and eligibility checklists are now widespread in scientific venues \cite{page2021prisma,equator}, their use risks reducing scientific judgment to mechanical rubric‑following \cite{logullo2020checklists}. This is problematic because real‑world studies often exhibit idiosyncratic or unanticipated issues that cannot be fully specified a priori, a limitation explicitly acknowledged in established risk‑of‑bias frameworks \cite{higgins2011rob}. A core component of scientific reasoning is therefore the ability to autonomously identify critical reasons for excluding a study from downstream analysis, without reliance on predefined schemas.

To evaluate this ability, we construct a benchmark from 100 open-access systematic reviews, each centered on a focused evidence-based medicine question. These reviews document both included and excluded studies, along with human-written justifications for exclusion by a panel of domain experts. From each review, we extract a precise causal research question, and then examine the excluded studies.


Each benchmark question takes the form:
“A panel of evidence-based medicine experts determined that the study <DOI> does not provide appropriate evidence to address the following research question: <RESEARCH QUESTION>. What was their justification for excluding this study?”
This open-ended format encourages agents to independently surface the most epistemically salient exclusion rationale, without being cued to any particular checklist. We manually inspect the final set of items to ensure each question is answerable by a domain expert using only the research question and study design. The result is a benchmark that captures a broad notion of scientific appropriateness, blending methodological scrutiny with relevance assessment—both critical to trustworthy AI assistance in evidence-based science.

\subsubsection{Literature task construction}
With the exception of SourceQualQA, the tasks comprising the literature understanding subtasks were generated by contracted domain experts who hold or are in the process of obtaining PhDs in biology or related disciplines. We provide our experts a purpose-built web-based platform for capturing their work and managing task generation, review, and revision processes. We provided papers, figures, or tables for them to work from, as well as allowed submission of their own papers of choice to the corpus.

We provided detailed instructions for question writing within each category, and each question went through multiple rounds of review and revision by both the same group of contracted experts and internal experts in order to assemble a high-quality set of questions with the intended features and targeting the intended capabilities. Most importantly, experts were instructed to ensure that questions they wrote could only be answered with the respective source, including other information sources entirely but also other parts of the same source document, where a specific component was being assessed (e.g. questions in FigQA2 should only be answerable by the figure in question and not the main text of the paper.) They were similarly instructed to ensure that answers were not \emph{contradicted} by any other source. Reviewers were asked to assess the same properties, as well as to ensure that the answers provided were correct and in the source and the information in the question was sufficient to find the correct source document for retrieval questions.

\subsection{Data Access: DbQA2}
\label{sec:data-access}

Modern biology research produces enormous amounts of data~\citep{li2014big,katz2022sra}, much of which is publicly hosted and available~\citep{clough2024geo,david2026ena}, and is to a large extent under-explored~\citep{mahi2019grein,nardini2015multiomic}. This presents an obvious opportunity area for deploying AI systems to leverage this under-explored and disconnected data for new discoveries. Measuring the ability to access and use this information is paramount. In \lbone, data access is assessed with DbQA, a set of tasks requiring access to specific entries in (sometimes multiple) biological databases to answer accurately. At the time of the \lbone publication, models had little access to search tools and DbQA was thus one of the most difficult subsets from \lbone ~\cite{labbench}, and remains difficult even for current models (see Figures~\ref{fig:lb1-lb2-comparison} and~\ref{fig:model-comparison}). Rather than simply convert DbQA into open-response questions, the approach to data access evaluation in \lbtwo is more opinionated about the specific data sources being measured. A larger number of data sources are targeted, and access to more specific and esoteric information within each is measured in order to more specifically measure true data access and avoid training knowledge contaminating results.

\subsubsection{DbQA2 construction}
To construct DbQA2, expert opinion was combined with an analytical approach to select the data sources to assess. An LLM-based analysis of ~1,000 recent papers posted to biorXiv~\cite{biorxiv_cshl_launch_2013} was used to determine where their input and output data was sourced or uploaded. The top ranked (by representation-fraction in the set of analyzed papers) data sources from this set were then manually curated, and additional sources were added based on input among internal scientific experts, yielding 43 sources. These were then provided to both internal domain experts and our group of contracted domain experts to draft source-specific questions. Questions went through multiple rounds of review and revision to arrive at the final set.

\subsection{Molecular Biology}
\label{sec:molecular-biology}

SeqQA was the primary molecular biology component of \lbone, consisting of a variety of formulaic tasks requiring understanding and manipulating biological sequences. It was targeted for evaluating specific capabilities benefiting in many cases from specialized tool use, such as DNA manipulation software, thus making it a very difficult set of tasks overall at the time of publication~\cite{labbench}.

For SeqQA2, a similar approach is used, but difficulty and realism are increased by (i) introducing several new subcategories of tasks and (ii) introducing the need to read sequences from files or retrieve them, in addition to providing them in-line. Where possible within a given subcategory, SeqQA2 tasks can be set to \emph{either} provide any input sequences in-line with the task prompt (similar to \lbone SeqQA), require reading a provided file, or require retrieval from a source of choice. For several subcategories, such as designing primers for PCR, there are inherently many possible valid correct designs. We thus created custom \emph{verifier functions} for each subcategory to carry out evaluation by e.g., performing an \textit{in silico} PCR with the model-designed primers.

\subsubsection{SeqQA2 construction}
SeqQA2 is constructed largely programmatically: Each of the 20 subtask types was initially developed as a question template by an in-house biologist, and then formulaically expanded to 20 variant questions each using alternative input sequences or data, either designed or as real gene sequences.

\begin{figure*}[t]
  \centering
  \includegraphics[width=\textwidth]{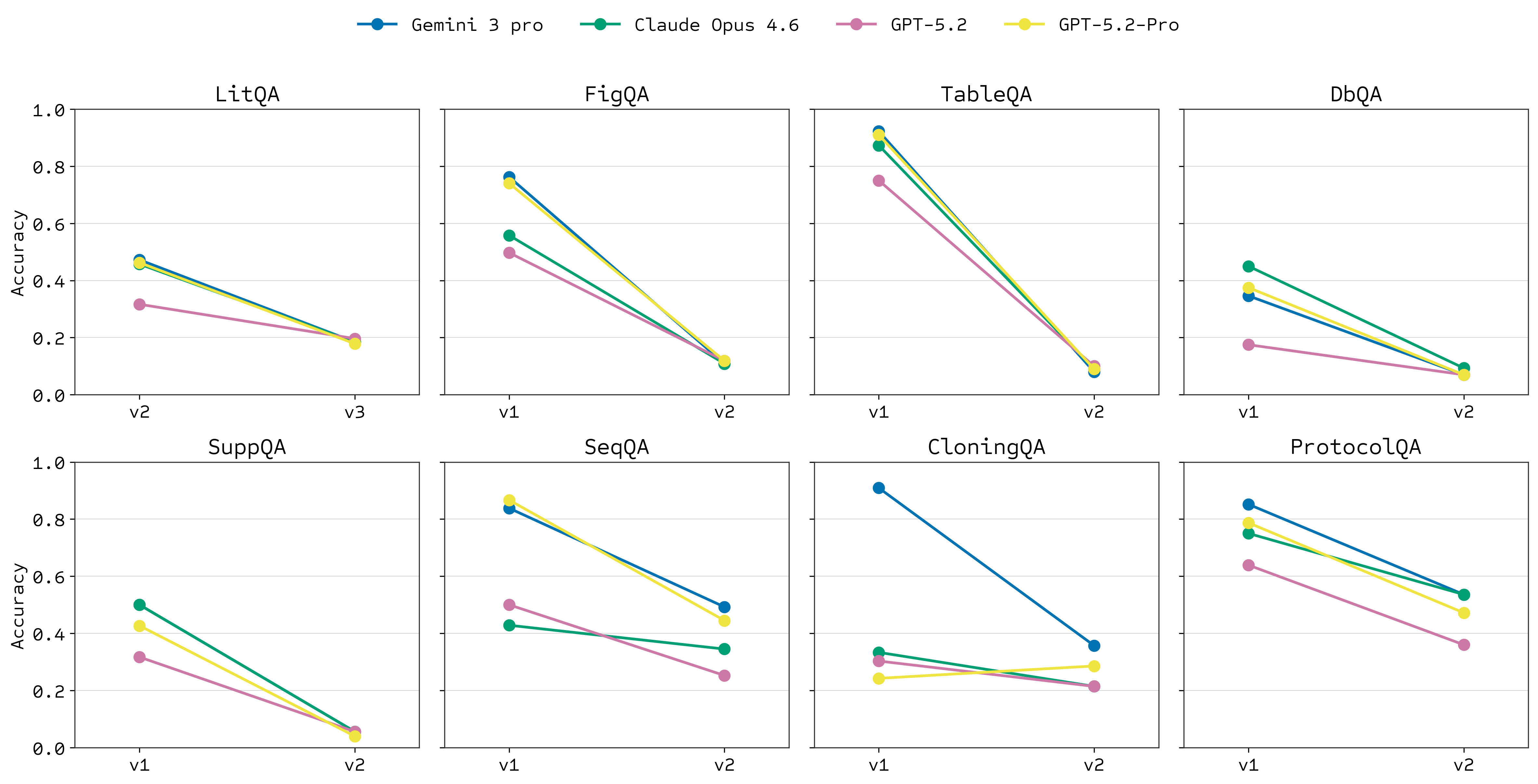}
  \caption{Accuracy score comparison between \lbone and \lbtwo for each of the high-level task families covered by both benchmarks.}
  \label{fig:lb1-lb2-comparison}
\end{figure*}

\subsection{Protocol Troubleshooting}
\label{sec:protocol-troubleshooting}

ProtocolQA from \lbone was designed as an early measure of model ability to troubleshoot error-containing lab protocols. At release, ProtocolQA was very difficult in MCQ form~\cite{labbench}, though labs like OpenAI additionally created their own open-response versions~\cite{openai_o1_system_card}. ProtocolQA2 extends the same approach in the original ProtocolQA of using modified protocols with introduced errors. To increase difficulty and realism, emphasis is placed on longer, more complex protocols, as well as requiring direct identification of errors introduced into the published form (e.g., in a PDF) in some cases.

\subsubsection{ProtocolQA Construction} ProtocolQA2 tasks were constructed by contracted domain experts. Protocols are pre-obtained both from public sources (protocols.io~\cite{teytelman2016protocolsio} and STAR Methods~\cite{cellpress2016star_methods}) and via a mechanism for experts to submit their own protocols, some of which are unpublished. Experts are instructed to introduce an error into the protocol (such as changing an incubation temperature) such that the error would cause an unambiguous negative result. They are then asked to formulate a question based on the error, posing a hypothetical scenario and result, with a prompt to identify an error that would cause the provided result. A prompt suffix also requests that the model return only a single important error, to avoid the default tendency to suggest several possible changes to the protocol.

\subsection{Experiment Planning}
\label{sec:experiment-planning}

The CloningScenarios subcategory from \lbone is a very difficult set of tasks that presented real-world molecular cloning scenarios encountered in real project contexts, with associated questions requiring thorough understanding of how molecular cloning experiments are planned, carried out, and interpreted. Many of the questions benefited from specialized tool use, though they were still multiple-choice questions targeting a narrow aspect of each scenario. Performance on the original CloningScenario task set has steadily risen to surpass human level~\cite{airiskmonitor2025q3report,anthropic_opus45_system_card}, though it bears reiterating the caveats of human evaluation on this subcategory outlined in the original paper~\cite{labbench}.

For \lbtwo, this concept is taken several steps further with CloningQA. These tasks require complete end-to-end design of a molecular cloning protocol, including design or selection of any necessary DNA or enzyme reagents and all protocol steps to complete the requested cloning scenario. The prompted scenarios range in difficulty, and are distributed between traditional restriction-ligation cloning, Gibson assembly, and Golden Gate cloning methods. For evaluation, reagent components and protocol steps are output in a simple domain-specific language format that can be parsed for individual component design verification and validated end-to-end using custom verifier functions \textit{in silico}.

\section{Results}
\label{sec:results}

To demonstrate the state of current frontier language models' ability to participate in foundational research activities, we evaluate \lbtwo performance for several frontier models. We report results both for base models (i.e. without tools) as well as for the models augmented with available tools, primarily web search and code execution.

\begin{figure*}[t]
  \centering
  \includegraphics[width=\textwidth]{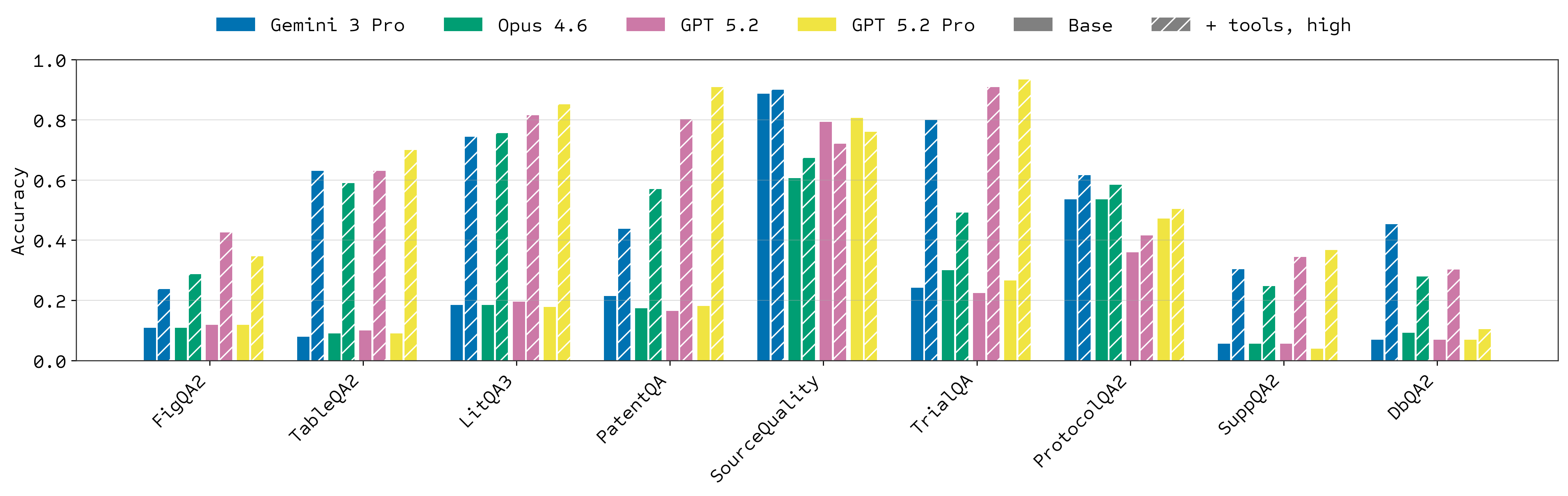}
  \caption{Performance comparison of frontier language models on \lbtwo broad task families. Results for base models (without tools) and with web search and code execution tools are indicated with solid or hashed bars, respectively.}
  \label{fig:model-comparison}
\end{figure*}

\subsection{\lbtwo is more difficult than \lbone}
We compared \lbtwo performance directly to \lbone performance at the broad task family level to demonstrate a meaningful increase in difficulty. Absolute scores are expectedly lower relative to \lbone. Figure~\ref{fig:lb1-lb2-comparison} summarizes accuracy on \lbone versus \lbtwo for corresponding families. Across all task families and models, we observe a consistent drop when moving from \lbone to \lbtwo, with model-specific differences in accuracy ranging from $-26\%$ to $-46\%$. This gap is driven by the higher-fidelity task framing in \lbtwo, including open-response answering, and more realistic contexts, such as retrieval- or file-dependent task framing.

\subsection{Frontier models variably benefit from tool augmentation}
Figure~\ref{fig:model-comparison} compares model performance across \lbtwo task families. Across several of the families, access to tools substantially benefits model performance. This is particularly evident for information retrieval-based tasks like LitQA3 and Patent/TrialQA. Accessing information in supplemental material also benefits meaningfully from tool-use, likely driven by web-search abilities, but is still well below retrieval from main text. We speculate that this is largely due to the wide array of file types used in supplemental material, including PDFs, Excel spreadsheets, CSVs, figures, etc.

Tool-use is not the complete solution either. Performance on FigQA2 (in the default \emph{retrieve} mode), SuppQA2, and DbQA2 in particular demonstrate that despite increased ability to reliably retrieve papers and access other web-based information with web search tools, unlocking the information contained within remains a significant performance bottleneck.

In contrast to retrieval over unstructured text, DbQA2 tasks require navigating specialized scientific databases, correctly accessing relevant record(s) and extracting precise information (often with nontrivial schema, identifiers, and multiple similarly named entities). DbQA2 remains one of the most challenging families in \lbtwo, even with tool augmentation (Figures~\ref{fig:lb1-lb2-comparison} and~\ref{fig:model-comparison}). Given these results and the substantial amount of high-value information contained in these sorts of data sources, this capability should become a key development focus.

\subsection{Models have impressive standalone visual understanding}
\begin{figure}[t]
  \centering
  \includegraphics[width=\columnwidth]{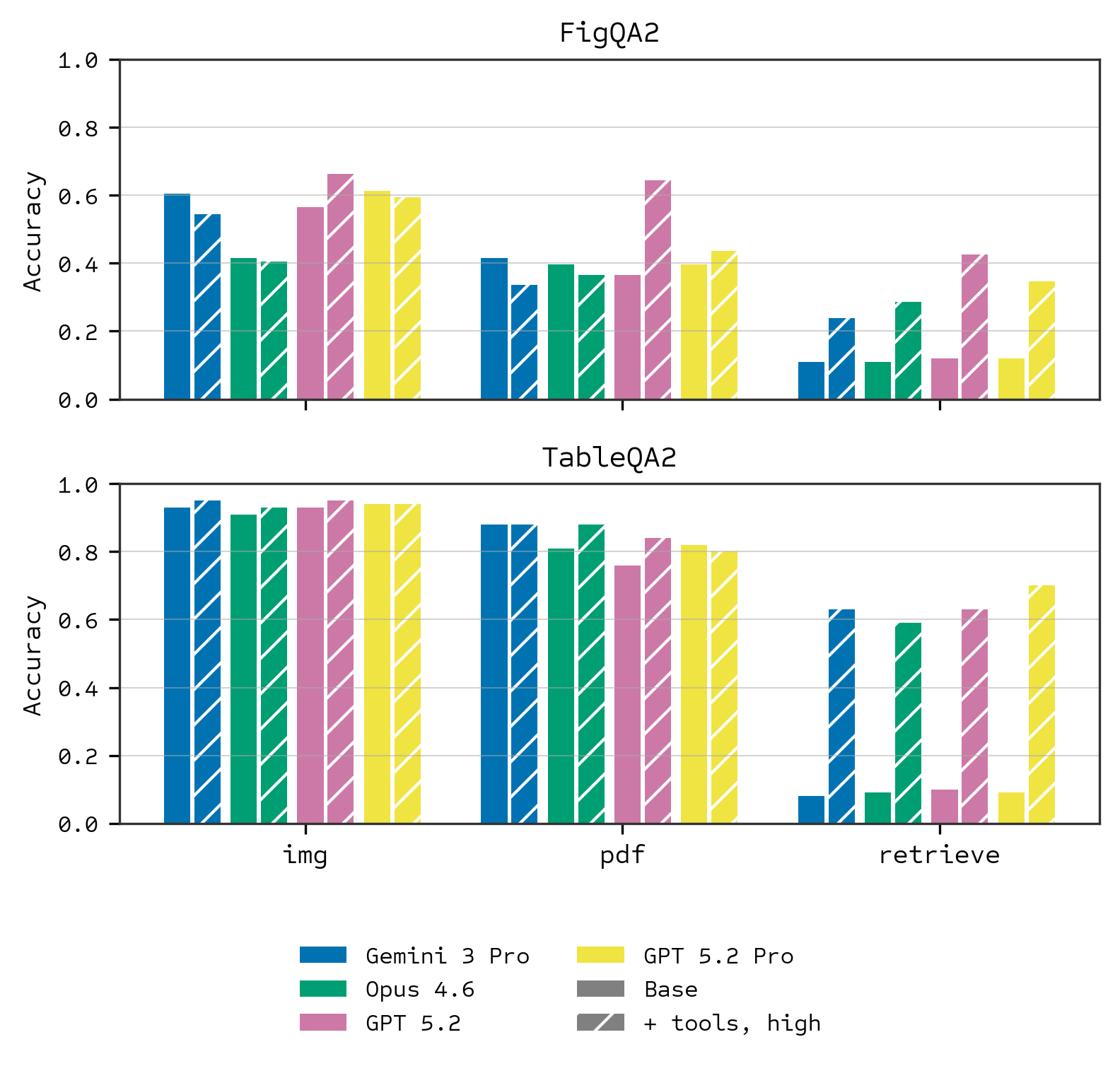}
  \caption{Performance on FigQA2 and TableQA2 across the three task modes (image-provided, paper-provided, and retrieval).}
  \label{fig:figqa-tableqa-modes}
\end{figure}

With the newly modified design of figure and table understanding tasks in \lbtwo, we can independently measure the ability to read figures/tables alone, in the context of the source paper, or layered on top of the need to retrieve the source paper. Figure~\ref{fig:figqa-tableqa-modes} shows the disparity in performance across the three modes for figure and table understanding. For both categories, performance unsurprisingly decreases moving from image-provided, to paper-provided, to retrieval modes. Though it is impressive the degree to which all frontier models can now understand figures and especially tables put in front of them, there is still a substantial gap to address when it comes to retrieving appropriate papers and understanding figures within them.

\subsection{Input modality is a key driver of molecular biology performance}
\begin{figure*}[t]
  \centering
  \includegraphics[width=\textwidth]{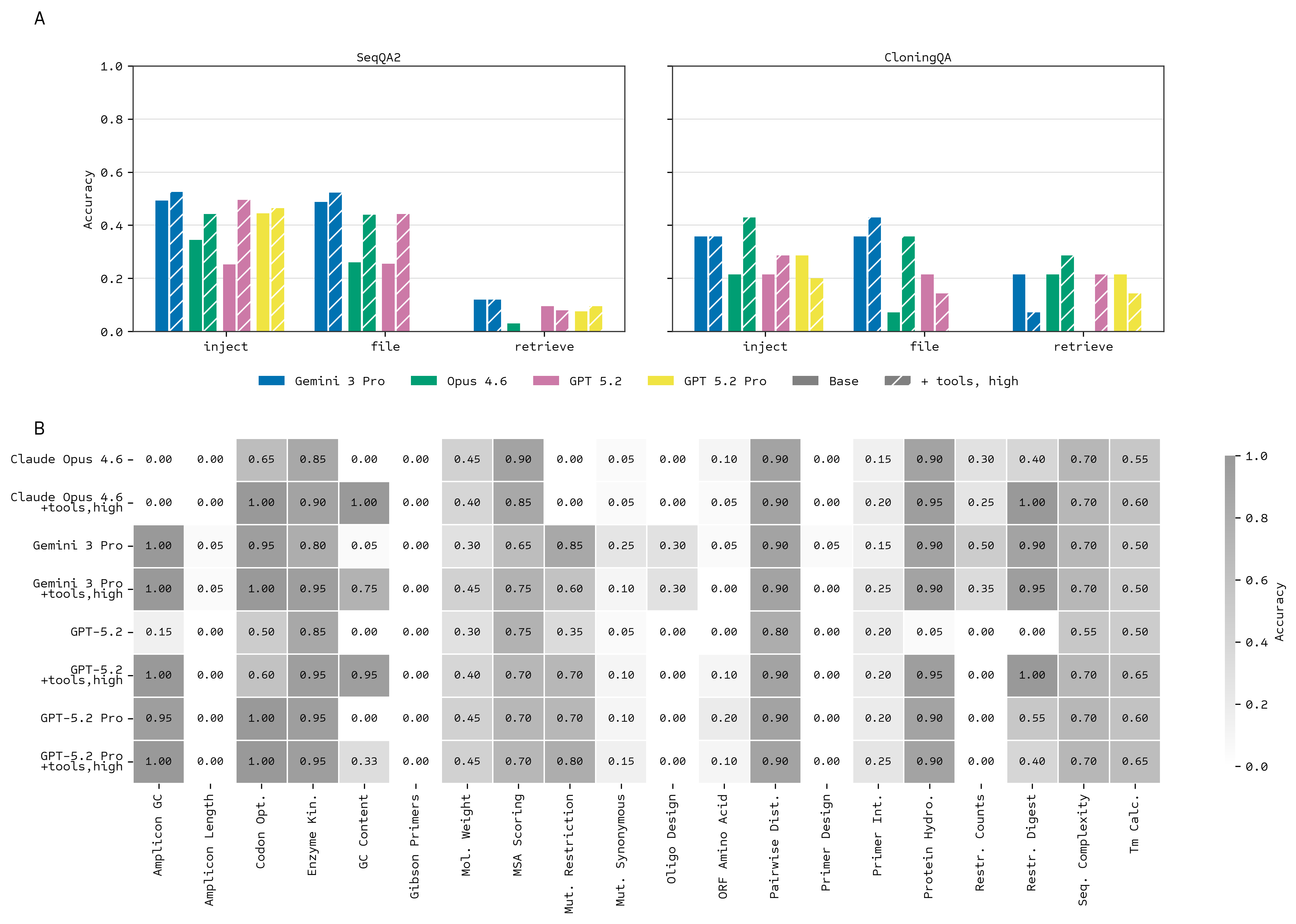}
  \caption{SeqQA2 and CloningQA performance. (A) Overall performance breakdown by sequence-input modality (inline, file, or retrieval). Note that GPT 5.2 Pro does not accept appropriate file types with the Response API, artificially deflating \texttt{file} mode performance. (B) Heatmap of SeqQA2 performance by subcategory across models. These results are in the default \texttt{inject} mode}.
  \label{fig:seqqa2-cloning}
\end{figure*}

Many of the subtasks in the SeqQA2 and CloningQA categories can be run in separate modes where necessary input context (i.e. a molecular sequence) is provided via direct prompt injection or file upload, or requires retrieval. Figure~\ref{fig:seqqa2-cloning}A breaks down performance by how this context is delivered. In-line prompting generally results in marginally better performance, though for models that allow it, file access is nearly equivalent for SeqQA2. Notably, file-based CloningQA tasks actually appear easier than injected tasks for Gemini 3 Pro and Opus 4.5 when augmented with tools, likely due to the long input sequences being dealt with in these tasks (often 3,000 base pairs or longer). As expected, tool-use (likely code execution in this case) also substantially increases SeqQA2 and CloningQA performance - actually acting as a `great equalizer' across models on these tasks. In all cases, and in line with the previously discussed DbQA2 performance, the retrieval-based variants show very poor performance, highlighting the substantial limitations models have for accessing specific data sources, even with the latest tools.

The SeqQA subcategory heatmap (Figure~\ref{fig:seqqa2-cloning}B) shows that performance is highly non-uniform across specific skills. Subtasks that require identifying or manipulating specific subsequences from larger sequence context (e.g. primer design, AA identification) show the lowest performance across all models. More global sequence operations (assessing sequence complexity, scoring sequence alignments) score more highly for all models, whereas some calculation-reliant tasks (GC content, enzyme kinetics) are benefit especially from tool use. This suggests that aggregate SeqQA2 scores can mask specific bottleneck skills, and should motivate subcategory-targeted development and evaluation.

\section{Discussion}
\label{sec:discussion}

\lbtwo is motivated by the simple premise that to be useful in real scientific workflows, models must reliably perform real-world research tasks. The results presented here support this motivation: despite clear progress on \lbone-evaluated task types, the more realistic framings in \lbtwo exposes gaps that are directly relevant to practical benefit.

\subsection{What \lbtwo reveals about current models}
\paragraph{Retrieval and localization abilities are still lacking.}
The largest performance drops arise when models must (i) identify the correct source, and then (ii) localize a specific figure/table/supplemental information within a long document. This suggests that improvements in reasoning alone are insufficient: robust document-grounded work requires better retrieval and better navigation over heterogeneous scientific documents and artifacts. Accessing non-text based information in specialized databases remains a substantial gap in capability, and one that must be addressed to unlock high-value AI research applications.

\paragraph{Faithful handling of exact inputs is still fragile.}
SeqQA2 and CloningQA highlight a different failure mode: even when the required operation is conceptually straightforward, correctness depends on exact string-level fidelity and using tools correctly. This is a well-known error source, and human experts have built many purpose-built tools to deal with things like faithful DNA sequence manipulation within complex protocols. Giving models access to these sorts of tools is an obvious parallel that should only benefit them when dealing with this sort of exacting task.

\paragraph{Scientific discernment is not the same as rubric compliance.}
The new SourceQuality family is designed to test whether an agent can surface the most epistemically salient reason a study is inappropriate for a research question. This capability is central for trustworthy evidence synthesis and is only partially addressed by checklist-style approaches; models must combine relevance assessment with methodological scrutiny.

\subsection{Implications for building research systems}
The broad results we find for frontier models on \lbtwo suggest several priorities for continued model and agent development:
\begin{itemize}[leftmargin=*]
  \item \textbf{Retrieval robustness:} agents should be robust to uncertainty around retrieval, and have built in mechanisms for iterative search and retrieval to ensure sufficient ``exploration'' of retrieval space.
  \item \textbf{Document-grounded extraction:} more effective parsing of PDFs and supplements, especially multi-modal artifacts like figures, require continue development.
  \item \textbf{Specialized data access:} DbQA2 highlights that robust interaction with scientific databases remains a key missing capability. Building agents that can reliably navigate database interfaces/APIs and extract data within is an especially important direction.
  \item \textbf{Dedicated sequence tools:} SeqQA2 and CloningQA show that fragile handling of long, exact sequences remains a major limitation for real-world deployment even in relatively ``simple'' contexts. Providing purpose-built sequence manipulation tools can substantially improve reliability.
\end{itemize}

\subsection{Limitations and future work}
While \lbtwo increases fidelity, it still simplifies many aspects of real research (e.g., wet-lab execution, ambiguous goals, and long-horizon iteration across days or weeks). Evaluation also still necessarily depends on the reliability of verifiers and grading. Real research tasks can have ambiguous hard- to impossible-to-verify outcomes, and are also compositions of the sorts of granular tasks comprising \lbtwo. Building reliable evaluation schemes and representative tasks to address these sort of abilities will be key to further AI Scientist developments. Further, as with \lbone, \lbtwo does not succeed at or attempt to comprehensively cover all of biological research capabilities. It is necessarily scoped to a set of abilities that are easily measurable yet fundamental to the research process as a whole.

Looking forward, we see several extensions as high value: (i) richer long-horizon task compositions that span multiple task types (e.g. literature search $\rightarrow$ data access \& analysis $\rightarrow$ protocol design $\rightarrow$ wet-lab experiments $\rightarrow$ result interpretation), (ii) methods for evaluating tasks with ambiguous or non-deterministic outcomes, e.g., multiple valid plans or partially correct approaches, and (iii) more focused evaluations in specific scientific sub-domains to better isolate capabilities and target high-impact, high-value application areas like drug development. We anticipate rapidly growing the LABBench family of benchmarks to increasingly cover evaluation in these areas.

\section*{Competing Interests}

JLS is an advisor to ForensisGroup Inc. and a scientific consultant at Anthropic Inc. During the course of this project, JLS was a scientific consultant at Edison Scientific, Inc. ASV is a scientific consultant at Edison Scientific, Inc.

\section*{Acknowledgments}

We thank our expert contractors for their extensive work in drafting, revising, and validating benchmark tasks across \lbtwo, and for their careful attention to source fidelity and answerability throughout the multi-round task generation process.

\bibliographystyle{icml2026}
\bibliography{references}

\newpage
\appendix
\onecolumn



\end{document}